\documentclass[acmlarge]{acmart}
\usepackage{booktabs}
\usepackage{hyperref}
\usepackage{url}
\usepackage{color}
\usepackage{graphicx}
\usepackage{caption}
\usepackage{subcaption}
\usepackage{wrapfig}
\usepackage{comment}
\usepackage{multirow}
\usepackage{rotating}
\usepackage{arydshln} 
\usepackage{xspace}

\makeatletter
\DeclareRobustCommand\onedot{\futurelet\@let@token\@onedot}
\def\@onedot{\ifx\@let@token.\else.\null\fi\xspace}

\def\eg{\emph{e.g}\onedot} 
\def\ie{\emph{i.e}\onedot} 
\def\cf{\emph{cf}\onedot}

\def\etal{\emph{et al}\onedot}
\makeatother

\def\BibTeX{{\rm B\kern-.05em{\sc i\kern-.025em b}\kern-.08emT\kern-.1667em\lower.7ex\hbox{E}\kern-.125emX}}
\copyrightyear{}
\acmYear{}
\setcopyright{rightsretained} 
\acmConference[Under review]{}{do not distribute}{}
\acmBooktitle{xxx}

\usepackage{etoolbox}
\patchcmd{\quote}{\rightmargin}{\leftmargin 3em \rightmargin}{}{}

\definecolor{dodgerblue}{rgb}{0.12, 0.56, 1.0}
\definecolor{skyblue}{rgb}{0.53, 0.81, 0.92}
\definecolor{limegreen}{rgb}{0.2, 0.8, 0.2}
\definecolor{greenmatplotlib}{rgb}{0.0, 0.5019607843137255, 0.0}
\definecolor{peru}{rgb}{0.8, 0.52, 0.25}
\definecolor{blueviolet}{rgb}{0.54, 0.17, 0.89}
\definecolor{chartreuse(traditional)}{rgb}{0.87, 1.0, 0.0}
\definecolor{chartreuse(web)}{rgb}{0.5, 1.0, 0.0}
\newcommand{\cSAT}{{\color{green}{SAT}}\xspace}
\newcommand{\cPD}{{\color{greenmatplotlib}{PD}}\xspace}
\newcommand{\cgSHAP}{{\color{orange}{gSHAP}}\xspace}
\newcommand{\gamfive}{{\color{dodgerblue}{SAT-5}}}
\newcommand{\gamtwo}{{\color{skyblue}{SAT-2}}}
\newcommand{\dtfive}{{\color{greenmatplotlib}{DT-4}}}
\newcommand{\dttwo}{{\color{limegreen}{DT-2}}}
\newcommand{\rulesfive}{{\color{blueviolet}{RULES-5}}}
\newcommand{\sparsetwo}{{\color{peru}{SPARSE-2}}}

\begin{document}
\title[Considerations When Learning Additive Explanations for Black-Box Models]{Considerations When Learning Additive Explanations for Black-Box Models} 
\author{Sarah Tan
}
\affiliation{%
  \institution{Facebook, Menlo Park, USA. Corresponding author. \url{ht395@cornell.edu}.}
}
\thanks{* This work started before the authors joined Facebook.}

\author{Giles Hooker}
\affiliation{%
  \institution{University of California, Berkeley, Berkeley, USA}
}

\author{Paul Koch}
\affiliation{%
  \institution{Microsoft Research, Seattle, USA}
}

\author{Albert Gordo
}
\affiliation{%
  \institution{Facebook, Menlo Park, USA}
}

\author{Rich Caruana}
\affiliation{%
  \institution{Microsoft Research, Seattle, USA}
}

\renewcommand{\shortauthors}{Tan et al.}

\begin{abstract}
Many methods to explain black-box models, whether local or global, are additive. In this paper, we study global additive explanations for non-additive models, focusing on four explanation methods: partial dependence, Shapley explanations adapted to a global setting, distilled additive explanations, and gradient-based explanations. 
We show that different explanation methods characterize non-additive components in a black-box model's prediction function in different ways. We use the concepts of main and total effects to anchor additive explanations, and quantitatively evaluate additive and non-additive explanations. Even though distilled explanations are generally the most accurate additive explanations, non-additive explanations such as tree explanations that explicitly model non-additive components tend to be even more accurate. Despite this, our user study showed that machine learning practitioners were better able to leverage additive explanations for various tasks. These considerations should be taken into account when considering which explanation to trust and use to explain black-box models.

\end{abstract}


\maketitle

\section{Introduction}
\label{sec:chapter_03_gae:introduction}

Whether explicitly or implicitly, many model explanation methods, whether local or global, require characterizing the impact of a particular feature on the machine learning model's prediction function. Many of these explanations are additive, explaining model predictions by adding up the impact of different features. For example, Shapley additive explanations \cite{Lundberg2017unified}, a local feature importance method, consists of Shapley values that add up to the prediction for a particular point. LIME \cite{lime} fits a locally linear -- hence locally additive --  model in the neighborhood around the point, with the weights from the linear model describing how features impact the prediction. 

However, when a model's prediction function has non-additive components, there is no one unique additive explanation that can perfectly characterize it \cite{hastie1986generalized}. Even when models only have additive components, but different components are correlated  \cite{Amodio2014concurvity}, there is no one unique additive explanation. This has been demonstrated in several practical settings; Orlenko and Moore \cite{orlenko2021comparison} simulated genetic datasets with non-additive epistatic interactions of different complexity, finding that while feature importance measures, including SHAP, were able to identify genes with the largest independent main effects, different feature importance measures did not judge genes involved in non-additive interactions similarly. Lengerich et al. \cite{lengerich2020purifying} found on the MIMIC-III dataset that the feature-prediction relationship for blood urea nitrogen (BUN), a feature that participated in a large number of discovered interactions, was significantly altered after performing a procedure proposed in their paper to move main effects out of interaction terms. 

Hence, we are left with a conundrum -- when there exists multiple additive explanations that look different from each other, \textit{how do we know which additive explanation method to trust and use}?

We study this question from several angles in this paper. First, we show how several popular additive explanation methods -- partial dependence \cite{friedman2001pdp}, Shapley additive explanations \cite{Lundberg2017unified}, distilled additive explanations \cite{tan2018dc}, and gradient-based additive explanations -- differentially characterize main effects of features in the presence of non-additive interaction effects that involve these features, or correlation between features. We use the concepts of main and total effects (related to marginal plots \cite{hastie2009elements}), to anchor these additive explanations, and show how each of them distributes interaction effects in a different way. 
Secondly, we quantitatively compare different additive explanations on several regression and classification tasks, finding that distilled explanations are generally the most accurate additive explanations (unsurprisingly, since their training optimizes for faithfulness to the black-box model) and that non-additive explanations that explicitly model interactions tend to be the most accurate over all. 

Despite these considerations, our investigations in this paper find that non-additive explanations are not a panacea. In a user study we performed to evaluate the interpretability of additive explanations compared to non-additive explanations that may not suffer from these same considerations, we found that machine learning practitioners were able to leverage additive explanations for various tasks, neither trusting the explanation less due to perceived oversimplification (as in the case of linear explanations) or being overwhelmed by the complexity of the explanation (as in the case of large decision trees). 

The contributions of this paper are:
\begin{itemize}
    \item Proposing distilled additive explanations. While distilled tree-based explanations have been proposed in the literature  \cite{craven1995extracting}, additive models had not been tried as distilled students, and additive explanations were typically learned using permutation-based approaches such as partial dependence or Shapley values. Our paper presents insights uncovered when learning additive explanations using distillation. 
\item A comprehensive study of leading additive explanation methods, writing them in the same mathematical framework so they can be compared to each other, investigating how they perform under key settings that not all methods, when first proposed, considered (non-additive interactions between features, correlation between features). By studying these explanations not just empirically on real data, but also comparing them to the theoretically grounded main and total effects that can be calculated from synthetic functions, our analysis provides a fresh view into the differences between these explanation methods beyond simply fidelity and accuracy of an explanation.  
\item To the best of our knowledge there has not been any user studies with expert users (like ours) that compare different classes of global explanations, in particular global additive and global non-additive explanations; our paper is the first to compare the interpretability of explanations that are visually very different. 
\end{itemize}

The rest of the paper is organized as follows.  Section \ref{sec:rw} discusses related work. In Section \ref{sec:background} we review several popular explanation methods and show how they are additive. We consider how additive explanations are impacted by interaction effects and correlation in Section \ref{sec:interaction_and_correlation}, and provide results of empirical investigations of these phenomena in Section \ref{sec:experiments}. In Section \ref{sec:chapter_03_gae:userstudy} we describe results from the user study. Section \ref{sec:conclusions} concludes the paper.

\section{Related Work}
\label{sec:rw}
\paragraph{\textbf{Global explanations.}} Post-hoc explanations of black-box models can be roughly divided into local methods that explain predictions at individual points and global methods that explain overall model predictions \cite{doshivelez2017towards}. Examples of global explanations include using a second interpretable model to approximate the black-box model, often termed ``model distillation'' \cite{fu1994rule, craven1995extracting,sanchez2015towards, frosst2017soft,bastani2017interpreting,lakkaraju2019faithful,Ibrahim2019global}, analyzing intermediate representations \cite{bau2017network, mu2020compositional} or concepts \cite{kim2018tcav} encoded by the black-box model, prototype selection \cite{bien2011prototype,kim2016examples,tan2020tree}, counterfactual explanations \cite{rawal2020beyond} that summarize actions that can be taken to change the black-box model's predictions for an entire population, and feature importance measures \cite{breiman2001random, Lundberg2017unified, fisher2019all, covert2020understanding, williamson2020efficient, yan2020if}. We focus on global post-hoc explanations in this paper, and investigate how many of these explanation methods rely on approximating non-additive functions additively.

\paragraph{\textbf{Trustworthiness of explanations.}} One viewpoint in the community is that post-hoc explanations of black box models can be unreliable or misleading \cite{rudin2019stop}, and that inherently interpretable models should be favored. For example, Slack \etal showed that one can carefully craft black-box models that lead to innocuous explanations while still producing biased outputs \cite{slack2020fooling}. A known problem of additive explanations that affects their trustworthiness is their multiplicity. Main effects and interaction effects can freely move around while producing the same prediction \cite{lengerich2020purifying}. In addition, different training strategies may lead to significantly different or even contradictory models \cite{chang2021interpretable}. In this paper we empirically investigate this phenomenon on additive explanations.

\paragraph{\textbf{Evaluation of interpretability.}} There is no universal definition of interpretability \cite{doshivelez2017towards}; many recent papers evaluate interpretability in terms of how a human uses the model to perform downstream tasks. These studies are typically performed on non-expert humans (\eg Mechanical Turkers) \cite{lage2019human, poursabzi2021manipulating}; recently, there has been more work evaluating interpretability on expert users \cite{lakkaraju2019faithful, kaur2020interpreting, jesus2021can}. A contribution of this paper is a user study comparing global additive and non-additive explanations on expert users.

\section{Background}
\label{sec:background}
In this section, we review definitions and concepts that will be used throughout this paper.

\subsection{Global and Local Additive Explanations}
Let $\mathbf{X}$ be a $n$ by $p$ matrix consisting of feature values of points $\{x^j_i, i=1, \ldots, p, j = 1, \ldots, n\}$ where $p$ is the number of features and $n$ is the number of points. Let the vector $X^j = (x^j_1, \ldots, x^j_p)$ represent all feature values of the $j$th point, and let the vector $X_i = (x^1_i, \ldots, x^n_i)$ represent the $i$th feature value across all points. 

\begin{definition}[Global additive explanations]
Given $F$, the prediction function of a black-box model, global additive explanations $G$ decompose $F$, as a sum of effects of $p$ features: 
\begin{equation}
G(\mathbf{X}) = h_0 + \sum_{i=1}^p h_i(X_i) + \sum_{i=1}^p \sum_{i'=1}^p h_{ii'}(X_i, X_{i'}) + \ldots 
\label{eqn:global_additive}
\end{equation}
and we measure the accuracy of the representation $G(\bf{X})$ by  $||F(\textbf{X}) - G(\textbf{X})|| = \text{error}_G$. $h_0$ is a bias term, $h_i(X_i)$ is the ``main effect'' of feature $X_i$ on $G$ while $h_{ii'}(X_i, X_{i'})$ is a second-order ``interaction effect'' of features $X_i$ and $X_{i'}$  on F (second-order as it consists of two features). This definition of $G$ also accommodates higher-order interaction effects, \ie third-order and beyond. Such a decomposition is not unique and each of the methods examined below will give different results. We will refer to main effects and interactions in terms of algebraic groupings rather than how those groupings were derived.
\end{definition}

\begin{definition}[Local additive explanations]

Given $F(X^j)$, the prediction function of a black-box model on a single point $X^j = (x^j_1, \ldots, x^j_p)$, local additive explanations $\Gamma^j$ decompose $F(X^j)$, as a sum of effects of $p$ features: 
\begin{equation}
\Gamma^j(X^j) = \phi^j_0 + \sum_{i=1}^p \phi^j_i(x^j_i)
\label{eqn:local_additive}
\end{equation}
In Section \ref{sec:chapter_03_gae:shap} we show how global additive explanations can be constructed from multiple local additive explanations.
\end{definition}

\paragraph{\textbf{Visualization and interpretability.}} Global additive explanations can be visualized as in Figure \ref{fig:chapter_03_gae:gSHAP} where we produce one plot per feature, where each plot's x-axis is the domain of input feature $X_i$ and the y-axis is the feature's contribution to the prediction $h_i(X_i)$. Partial dependence plots \cite{friedman2001pdp} and Shapley dependence plots \cite{Lundberg2017unified} are visualized this way. As the number of features $p$ increases, there are more plots for a human to review and understand, which could decrease the interpretability of such explanations.

\subsection{Additive Explanations}
\label{sec:additive_explanations}
We now review several popular explanation methods and show how they can be written as additive equations. 

\subsubsection{Partial Dependence and Marginal Plots} The partial dependence \cite{friedman2001pdp} for feature $X_i$ is defined as:
\begin{equation*}
\label{eqn:partial_dependence}
q_i(X_i) = E_{X_C}[F(X_i, X_C)] = \int F(X_i, X_C) dP(X_C)\end{equation*} 
where $X_C$ is the set of all features excluding the $i$th feature. In practice, $q_i(X_i)$ is the average prediction from the data set when the value $X_i^j$ is replaced with the $X_i$ of interest. Specifically, for a specific value $z$:
\begin{equation*}\widehat{q_i}(X_i=z) = \frac{1}{n}\sum_{j=1}^n F(x_1^j, \ldots, x_{i-1}^j, x_i^j = z, x_{i+1}^j, \ldots, x_p^j) \end{equation*} 
Partial dependence plots of different features are often interpreted as being summed up as in Equation \ref{eqn:global_additive} with $h_i(X_i) = q_i(X_i)$ to construct a global additive explanation. However, to make this precise, the $q_i(X_i)$  need to be centered. This is because each $q_i(X_i)$ averages the values of $F$ and their sum will average to approximately $p$ times the average of $F$.   To center them, we subtract $\overline{q_i} = \frac{1}{|z|}\sum_{z}\widehat{q_i}(X_i=z)$, the average of $q_i(X_i)$ over all possible values of $z$, to obtain the global additive explanation:
\begin{equation}
PD(\mathbf{X}) = \hat{h}_0 + \sum_{i=1}^p \left\{ \widehat{q_i}(X_i)-\overline{q_i} \right\}
\end{equation}
where $\hat{h}_0 = \frac{1}{n} \sum_j F(X^j)$.  Partial dependence takes the expectation of the prediction function over the \textit{marginal} distribution of $X_C$, independent of the value of $X_i = z$. Alternatively, this expectation could also be taken over the \textit{conditional} distribution $X_C|X_i=z$. This is the marginal plot\footnote{The name is counter-intuitive as it is based on the conditional, not marginal distribution.} \cite{hastie2009elements, zhao2021causal, apley2020visualizing}, defined as: 
\begin{equation*}
\label{eqn:marginal_plot}
r(X_i) = E_{X_C|X_i}[F(X_i, X_C) | X_i] = \int F(X_i, X_C) P(X_C | X_i) dX_C \end{equation*} 
In practice, marginal plots are constructed by finding the collection of points for which $X_i = z$, $N(z) = \{ j: X_i^j = z \}$ and averaging over these for each $z$:
\begin{equation*}\widehat{r_i}(X_i=z) = \frac{1}{|N(z)|}\sum_{j \in N(z)} F(X^j) \end{equation*} 
Marginal plots of different features can also be centered and summed up to construct a global additive explanation: 
\begin{equation}
M(\mathbf{X}) = \hat{h}_0 + \sum_{i=1}^p \left\{ \widehat{r_i}(X_i)-\overline{r_i} \right\}
\end{equation}
where $\overline{r_i}$ is constructed using a similar centering procedure on $r_i(X_i)$ as was performed on partial dependence plots. Note that while partial dependence plots will recover components of an underlying additive function, marginal plots will only do so if the features are all independent.

\begin{figure}[t!]
\centering
\includegraphics[width=0.5\linewidth]{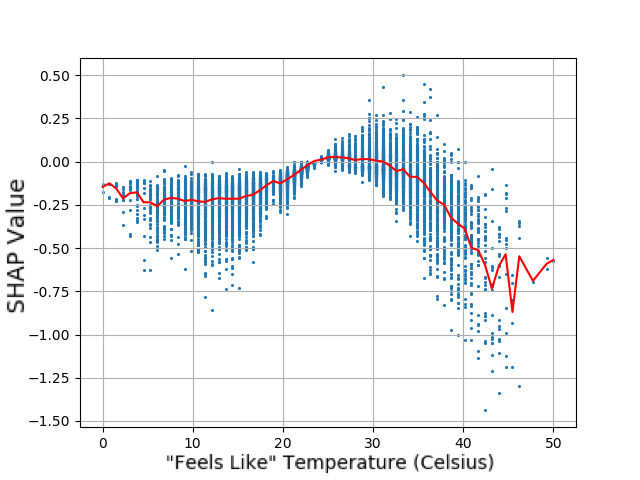}
\caption[From local Shapley explanations to gSHAP]{From \textcolor{blue}{local Shapley explanations} to \textcolor{red}{gSHAP}, a global Shapley additive explanation for a particular feature created by aggregating local Shapley explanations.}
\label{fig:chapter_03_gae:gSHAP}
\end{figure}

\subsubsection{Shapley Additive Explanations} 
\label{sec:chapter_03_gae:shap}

The Shapley local additive explanation \cite{Lundberg2017unified} for a point $X^j$ consists of $\phi_i^j$ that are defined to sum  to $F(X^j)$ up as in Equation \ref{eqn:local_additive}. To achieve this, we need to define prediction functions $F_S(X_S)$ for each subset $S$ of the features. \cite{Lundberg2017unified} defines $F_S(X_S) = \int F(X_S,X_C) P(X_C) dX_C$ as in partial dependence plots, but alternative formulations are also possible. The Shapley value for feature $X_i$ at point $X^j$ is then given by:
\begin{equation}
\label{eqn:shap}
\phi_i^j(x_i^j) = \sum_{S \subseteq \{x_1^j, \ldots, x_p^j\} \setminus x_i^j} \frac{|S|!(p-|S|-1)!}{p!} [F_{S \cup \{i\}}(X^j_{S \cup \{i\}}) - F_S(X^j_S)]
\end{equation}
For example, each blue point in Figure \ref{fig:chapter_03_gae:gSHAP} is one such $\phi_i^j$ for feature $i =$ ``Temperature'' and a particular point $j$. In practice, Equation \ref{eqn:shap} is computed by approximation, using techniques that repeatedly query the black-box model such as Kernel SHAP \cite{Lundberg2017unified}, Shapley sampling values \cite{vstrumbelj2014explaining}. 

Note, importantly, that while $F(X) = \sum \phi_i^j(X_i^j)$ at each point $X^j$, this does not directly define an additive model in the sense of \eqref{eqn:global_additive} since $\phi_i^j(X_i^j)$ changes with features other than $X_i$. Here we aggregate local explanations that act on single points to construct a global explanation \cite{bhatt2020, setzu2021glocalx, van2019global}. For example, the red line in Figure \ref{fig:chapter_03_gae:gSHAP} was created using the basic approach of feature-wise sample mean \cite{bhatt2020}, averaging $\phi_\text{Temperature}^j$ at each unique temperature value $z$ to construct a global additive explanation that we call \textbf{gSHAP} and define as:
\begin{equation}
\label{eqn:gshap}
gSHAP(\mathbf{X}) = \hat{h}_0 + \sum_{i=1}^p \frac{1}{n} \displaystyle \sum_{j=1}^n \phi_i^j(x_i^j)
\end{equation}
While gSHAP is not the only way to construct a global explanation from Shapley local explanations, it is intuitive and easy to implement on top of Shapley local additive explanations, hence we experiment with it in this paper. Another example of global Shapley values is SAGE \cite{covert2020understanding} which decomposes a model's accuracy function, but not prediction function, into Shapley values of each feature.

\subsubsection{Distilled Additive Explanations} Different from the previous methods that construct explanations by repeatedly querying the black-box model, this method treats the black-box model as a ``teacher'' model and uses model distillation techniques \cite{bucilua2006compression,ba2014deep,hinton2014distilling} to train a white-box model (``student'' model) that approximates and explains $F$. Different white-box model classes can be as the student model, such as decision trees \cite{craven1995extracting, frosst2017soft}, generalized additive models \cite{tan2018dc}, linear models, etc. 
If an additive model is used, the resulting explanation can be written as: 
\begin{equation}
AM(\mathbf{X}) = \hat{h}_0 + \sum_{i=1}^p \hat{h}_i(X_i)
\label{eq:distilled_additive}
\end{equation}
where the $\hat{h}_i$  are obtained by minimizing the mean-squared error between the teacher model's prediction function $F$ and student model's explanation $AM$ on training samples:

\begin{align*}
\hat{h}_0, \hat{h}_1, \ldots, \hat{h}_p & = \text{argmin } L(h_0, h_1, \ldots, h_p) = \text{argmin } \frac{1}{n} \sum_{j=1}^n \|F(X^j) - G(X^j) \|^2_2 \\
& = \text{argmin } \frac{1}{n} \sum_{j=1}^n \left|F(X^j) - \left(h_0 + \sum_{i=1}^p h_i(x^j_i)\right) \right|^2_2
\end{align*}
In this paper we consider two types of additive models: spline-based \cite{wood2006generalized} and tree-based \cite{lou2012intelligible} whose resulting explanations we call \textbf{Student Additive Splines} (SAS) and \textbf{Student Bagged Additive Boosted Trees} (SAT). For SAS, $\hat{h}_i(X_i)$ is a cubic regression spline function of feature $X_i$ implemented using the \texttt{mgcv} library \cite{mgcv}. For SAT, $\hat{h}_i(X_i)$ is a bagged tree function of feature $X_i$ implemented using the \texttt{InterpretML} \cite{nori2019interpretml} package. Pairwise interactions can also be modeled explicitly \cite{lou2013accurate}, in the form of $h_{ii'}(x^j_i,x^j_{i'})$ added to Equation \ref{eq:distilled_additive}. We call this resulting explanation \textbf{SAT + pairs}. 

\subsubsection{Gradient-Based Explanations} 
Finally we consider a gradient-based explanation that involves constructing an additive function $gGRAD$ through the first-order Taylor decomposition of $F$: 
\begin{equation}gGRAD(\mathbf{X}) = F(\textbf{0}) + \displaystyle \sum_{i=1}^p \frac{1}{n} \displaystyle \sum_{j=1}^n \frac{\partial{F(X^j)}}{\partial x_i^j}x_i^j
\end{equation}
This is related to the ``gradient*input'' method (\eg \cite{shrikumar2017learning}) used to generate saliency maps for images. While there are several other gradient-based feature attribution methods (\cite{simonyan2014deep,simonyan2015very,bach2015pixel,shrikumar2017learning}; also see \cite{montavon2017methods} or \cite{ancona2018towards} for a review), the ``gradient*input'' method is simple, intuitive, and studied a lot in the literature \cite{ancona2018towards, adebayo2019sanity}, hence we experiment with it. 

\section{Non-Additive Interaction Effects and Correlation}
\label{sec:interaction_and_correlation}
We now consider how additive explanations are impacted by non-additive interaction effects and correlation between features. For ease of exposition, we will assume that the black-box model's prediction function $F$ can be written as: 
\begin{equation*}
F(\mathbf{X}) = e_0 + \sum_{i=1}^p e_i(X_i) + \sum_{i=1}^p \sum_{i'=1}^p e_{ii'}(X_i, X_{i'}) + \text{error}
\label{eqn:prediction_function}
\end{equation*}
by decomposing $F$ into successive additive components of increasing complexity and using the error term to capture any remaining effects. Extending to this 2nd-order model allows us to explore the effect of non-additivity on additive explanations. 

\paragraph{\textbf{Main effect.}} A simple way to report the impact of feature $X_i$ on prediction function $F$ is to simply ignore (i) any interaction effects involving $X_i$; (ii) any correlation between $X_i$ and other features. This results in a measure of the impact of feature $X_i$ on $F$ that use only the main effect of $X_i$:
\begin{equation}\label{eqn:main_effects} M(X_i) = \texttt{func}[e_i(X_i)] \quad \quad \quad M(X_{i'}) = \texttt{func}[e_{i'}(X_{i'})] \end{equation} 
For example, when 
\texttt{func} is the variance function, this is the main effect Sobol index \cite{sobol1990sensitivity, owen2014sobol}. While main effect measures undercounts the impact of $X_i$ on $F$ in case of (i) 
this sidesteps the challenging issue of determining how to allocate the interaction effect $e_{ii'}(X_{ii'})$ between features $X_i$ and $X_{i'}$.

\paragraph{\textbf{Total effect.}} Another simple way to report the impact of feature $X_i$ on $F$ is to attribute an interaction effect to each feature that was part of that interaction effect. This results in a measure of the impact of feature $X_i$ on $F$ that use the main effect of $X_i$ and all interaction effects containing $X_i$:
\begin{equation}
\label{eqn:total_effects}
T(X_i) = \texttt{func}\{e_i(X_i) + \displaystyle \sum_{i'=1; i \neq i'}^p e_{ii'}(X_i, X_{i'})\} \quad \quad \quad
T(X_{i'}) = \texttt{func}\{e_{i'}(X_{i'}) + \displaystyle \sum_{i'=1; i \neq i'}^p e_{ii'}(X_i, X_{i'})\}
\end{equation}
For example, 
when \texttt{func} is the variance function, this is the total effect Sobol index \cite{sobol1990sensitivity, owen2014sobol}, and when \texttt{func} is the expectation function, depending on the domain that the expectation is taken over, this is either the marginal effect (the best least squares approximation to $F$ using only feature $X_i$ \cite{hastie2009elements}, also called marginal plots in \cite{apley2020visualizing}) or partial dependence plots \cite{friedman2001pdp}). 
In cases where \texttt{func} satisfies countable additivity (\eg the expectation function, where the expectation of the sum equals the sum of the expectations), marginal plots can also be produced for individual interactions or for groups of interactions.
While total effect measures overcount the impact of $X_i$ on $F$ in both cases (i) and (ii), this also sidesteps the challenging issue of determining how to allocate the interaction effect $e_{ii'}(X_{ii'})$ between features $X_i$ and $X_{i'}$, albeit in a different way, by allocating it to \textit{all} features that make up that interaction. 

\paragraph{\textbf{Distributed effect.}} It is clear that neither main nor total effects are the ``right" way to allocate an interaction effect between features, though they can still be useful to anchor other additive explanations. In between them are approaches that split up an interaction effect between different features -- For example, distillation approaches that train additive student models to mimic the prediction functions of non-additive black-box models \cite{tan2017detecting} express non-additive components as a best-fit additive approximation added to additive components. Shapley additive explanations \cite{Lundberg2017unified} distribute an interaction effect equally among features that made up that interaction. The aim of this paper is to study these explanations in the presence of non-additive interaction effects and correlation between features.
  
\subsection{Example}
\label{sec:interactions_example}
To make this more concrete, we consider an example function for which we have its analytical form and can hence compute the main and total effects for each feature: \begin{equation}
\label{eqn:F_2}
F_2(\textbf{x}) = F_1(\textbf{x}) + x_1x_2 + |x_3|^{2|x_4|} + \sec(x_3x_5x_6)
\end{equation}
where
\begin{equation}
\label{eqn:F_1}
F_1(\textbf{x}) = 3x_1 + x_2^3 -\pi^{x_3} + \exp(-2x_4^2) + \frac{1}{2+|x_5|} + x_6 \log(|x_6|) 
+ \sqrt{2|x_7|}  + \max(0,x_7)  + x_8^4 + 2\cos(\pi x_8)    
\end{equation} 
This non-additive, highly nonlinear function consists of components from synthetic functions proposed by \cite{friedman2008predictive}, \cite{hooker2004discovering} and \cite{tsang2017detecting}. Like \cite{tsang2017detecting}, we set the domain of all features to be independent and Uniform[-1,1]. Like \cite{friedman2008predictive}, we add noise features ($x_9$ and $x_{10}$) to our samples that have no effect on $F_1(x)$. 

The \textbf{main effect} for feature $x_4$ is $\exp(-2x_4^2)$. The \textbf{total effect} (specifically, the marginal plot) of feature $x_4$ is $E\left[\exp(-2x_4^2) + |x_3|^{2|x_4|} | x_4\right] = \exp(-2x_4^2)+\int_{-1}^{1} |x_3|^{2|x_4|} dP(x_3|x_4) = \exp(-2x_4^2) + \frac{1}{2 |x_4| + 1}$. This contribution of the interaction term to the impact of feature $x_4$, $\frac{1}{2 |x_4| + 1}$, manifests as the upward pointing cusp (\textcolor{gray}{gray line}) in Figure \ref{fig:synthetic_2_explanations}, pointier and higher than main effect alone (\textbf{black line}). 

However, for features $x_1$ and $x_2$, their \textbf{total effects} are actually the same as their \textbf{main effects}, as $E[x_1 x_2 | x_2] = E[x_1 x_2 | x_1] = \left.\int_{-1}^1 x_1 x_2 dP(x_2|x_1) = \frac{x_1}{2} \frac{x_2^2}{2}\right\rvert_{-1}^1 = 0$, despite the presence of an interaction term $x_1 x_2$ in $F_2$. As we will see in Figure \ref{fig:synthetic_2_explanations}, in this case, all additive explanations were not visually different from the main effects only line (\textbf{black line}). Hence, in this paper, whenever the black-box model function is known, we will provide the main and total effects estimates to ground the additive explanations studied.

\section{Experiments}
\label{sec:experiments}
We now conduct empirical investigations of how non-additive interaction effects and correlation between features may impact additive explanations.

\paragraph{\textbf{Black box models:}}
We use multi-layer perceptrons (MLP) neural networks as our black box models, though other models such as gradient boosted trees or random forests could also be used. These models contain non-additive interaction effects, even for very shallow networks, which makes them good candidates for our problem. As is standard, we use ReLU non-linearities between the different layers of the model. We also experimented with using batch normalization \cite{ioffe2015batchnorm} between layers, but we did not observe consistent gains to justify their use. 

Models were trained using cross-entropy for classification problems, and root mean square error (RMSE) for regression ones. We divided the data in train, validation and test splits, where 70\% of the data was used for training, 15\% for validation, and 15\% for test, and used 5 train-validation folds.

We used random search to find the optimal hyperparameters (number of hidden units per layer, learning rate, weight decay, dropout probability, batch size, etc) based on average validation performance on the train-validation splits and multiple random Xavier initializations \cite{glorot2010understanding}. Following common practice we select powers of 2 for the number of hidden units and batch size, and we search using a logarithmic scale to find the optimal learning rate and weight decay.
We use the Adam optimizer \cite{Kingma2015adam} with default beta parameters, and early stopping based on validation loss. Training was performed in PyTorch.

We experimented with different depths (number of layers), and found that models of a given depth behaved in a consistent manner. In most cases, models with one hidden layer underfit compared to models with two layers, i.e., both their validation and training loss was higher. Models with three or four hidden layers systematically obtained almost-perfect training loss, but their performance on the validation and test sets was not as good, even when strong regularization through dropout and weight decay was applied.

Based on those results, we selected two black box models for our experiments throughout the paper: a 2-hidden layers model with 512 hidden units per layer (2H-512,512), which will be our main model, and a 1-hidden layer model with 8 units (1H-8), which will be our low-capacity model used to evaluate trade-offs between accuracy and interpretability.

\paragraph{\textbf{Quantitative evaluation of explanations:}}
 We will quantitatively evaluate explanations as if they were models, following Lundberg and Lee \cite{Lundberg2017unified} who suggested viewing an explanation of a model's prediction as a model itself. Specifically, we evaluate 
 accuracy (how well the explanation predicts the original label). 
 A similar evaluation was performed by \cite{kim2016examples} who used their explanations (prototypes) to classify test data. 

 To learn our explanations we follow a similar approach than to train our black box models: we perform 5 different train-validation data splits (70\% of the data for train, 15\% for validation and 15\% for test) and tune all necessary parameters aiming at maximizing the accuracy on the validation set.
 
 \begin{table}
\begin{tabular}{lccc}
\toprule
         & Model & SAT explanation (distilled) & PD explanation \\ \midrule
1H-8 model & 0.46 & 0.46 & 0.46\\
2H-512,512 model &  0.14 & 0.12 & 0.12\\ 
\bottomrule
\end{tabular}
\caption{RMSE of 1H and 2H models, SAT and PD explanations on $F_1$. The accuracy of a GAM model trained directly on the data is 0.02 RMSE.}
\label{tab:f1accuracy}
\end{table}

\subsection{Preliminaries: do explanation explain the black-box model or the original data?} 
\label{sec:explanations_model}

\paragraph{\textbf{Hypothesis:}}
Explanations explain the black-box model behavior, not the original data.

\paragraph{\textbf{Setup:}}
Given a model trained on data produced by a known formula (\eg $F_1$), we can compare an explanation of the model with the model trained directly on the data. When the explanation of the data is accurate, we can attribute the differences between the explanations to the black-box model behavior, and not to lack of fidelity of the explanations.
We use function $F_1$ introduced in Section \ref{sec:interaction_and_correlation}, that does not include any interactions between features. 
Although this function is easily modeled by additive models, neural nets struggle to model it accurately due to how they combine all features through each layer, forcing the introduction of combinations that do not exist in the original function.
We simulate 50,000 samples, and train two neural nets models we aim to explain, a 2H-512,512 and 1H-8, to predict $F_1$ from the ten features. The high capacity 2H-512,512 neural net obtained a test accuracy RMSE of $0.14$, while the low-capacity 1H-8 neural net obtained a test accuracy RMSE of $0.46$, more than 3x worse (\cf Table \ref{tab:f1accuracy}).
 Next, we train: (1) An additive model trained directly to predict the labels in the data. Since this model is very accurate (RMSE of 0.02), it will help us anchor the explanations.  (2) Explanations based on the same additive models, trained to mimic the output from the black-box neural nets. For this experiment, we focus on tree-based additive explanations, but the conclusions also apply to the other additive explanations. 
 (3) Explanations based on partial dependence (similar behavior is observed with Shapley additive explanations).

\paragraph{\textbf{Results:}} 

\begin{figure}
\centering
\begin{tabular}{ccccc}
\hspace*{-0.2in}
\begin{turn}{90}
\hspace{2cm}  \scriptsize{$ h_i(x_i)$}\end{turn}
\includegraphics[width=0.49\linewidth, clip, trim={0cm 1.25cm 0cm 0cm}]{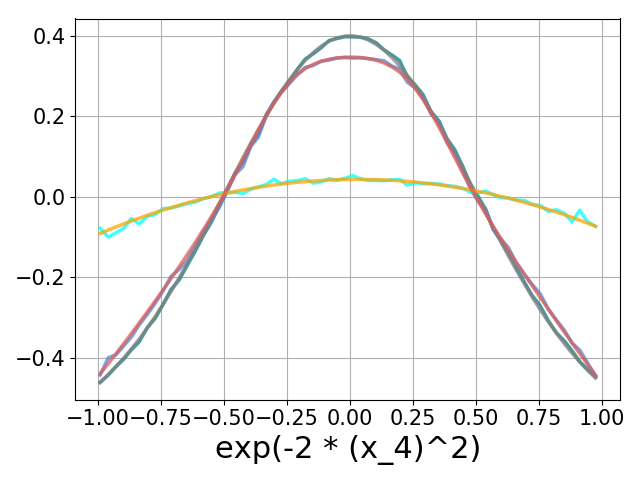} & \hspace*{-0.15in}
\includegraphics[width=0.49\linewidth, clip, trim={0cm 1.25cm 0cm 0cm}]{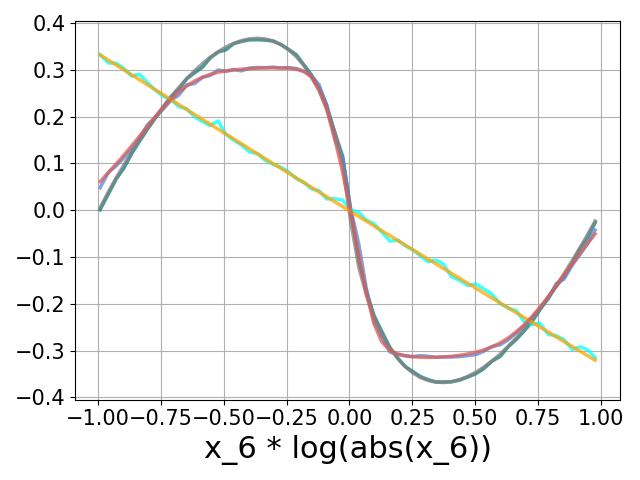} & 
\\
$x_4$ & $x_6$
\end{tabular}
\caption{For two features of function $F_1$: \textbf{Main effect}, \textcolor{green}{GAM trained on data}, \textcolor{dodgerblue}{SAT explanation of complex black-box model (2H-512,512 neural net)},
\textcolor{red}{PD explanation of the same 2H model},
\textcolor{cyan}{SAT of simpler black-box model (1H-8 neural net)},
and \textcolor{orange}{PD explanation of the same 1H model}.
}
\label{fig:chapter_03_gae:synthetic}
\vspace{-0.5cm}
\end{figure}

Figure \ref{fig:chapter_03_gae:synthetic} focuses on two features of $F_1$ ($x_4$ and $x_6$) and displays: (1) A plot of their analytical expression (in \textcolor{gray}{gray}). (2) Main effects of an additive model trained directly on the data (in \textcolor{green}{green}). (3) 
Distilled additive explanations of the two neural nets (in \textcolor{dodgerblue}{blue} and \textcolor{cyan}{cyan}). (4) Partial dependence plots for the two neural nets  (in \textcolor{red}{red} and \textcolor{orange}{orange}). Their accuracy is shown in Table \ref{tab:f1accuracy}.

We make the following observations: \begin{itemize}
    \item Additive models trained to directly predict labels in the data (in \textcolor{green}{green}) match almost-perfectly the  analytical expression (in \textcolor{gray}{gray}), showing that additive models have enough capacity to model additive functions without interactions. This models achieves a 0.02 RMSE, a very good value compared to the RMSE achieved by the neural nets (0.14 and 0.46).
\item The distilled additive explanation of the 2H-512,512 neural net (in \textcolor{dodgerblue}{blue}) largely matches the analytical expression (in \textcolor{gray}{gray}), but fails to match some complex parts of the shape, unlike additive models trained to directly predict labels in the data (in \textcolor{green}{green}). Quantitatively, this model achieves an RMSE of 0.12, comparable -- in fact, better, consistent with other observations regarding model distillation -- to the 2H-512,512 teacher, but worse than the 0.02 RMSE achieved by the additive model trained directly to predict the data labels.
Since the additive model has enough capacity to almost-perfectly model the data, we conclude the differences observed in the explanation of the 2H-512,512 neural net must be attributed to the neural net behavior.

\item The distilled additive explanation of the 1H-8 neural net (in \textcolor{cyan}{cyan}) looks notably different than the distilled additive explanation of the 2H-512,512 neural net (in \textcolor{dodgerblue}{blue}), and are also less similar to the analytical expression (in \textcolor{gray}{gray}). This is consistent with the RMSE error of the 1H model (0.46 RMSE) and the error of the student (also 0.46 RMSE). 
\item PD explanations are similar to the distilled explanations, both in terms of accuracy and in terms of shape. We clearly see the differences in explanations for the 1H and 2H models. 
\end{itemize}
The behavior on other features and other explanation methods is similar to the one observed with SAT and PD explanations on features $x_4$ and $x_6$.

\paragraph{\textbf{Implications:}} Since the additive model trained on original labels was able to almost-perfectly match the data (0.02 RMSE), and since all additive models were trained following the same protocols, the  differences observed on the distilled explanations can be safely attributed to the behavior of the neural net models. The same conclusion applies to PD, where we see clear differences between the explanations of the 1H and 2H models which are similar to the differences observed on the distilled explanations.

\subsection{How Do Different Additive Explanations Characterize Non-Additive Interaction Effects?}

\paragraph{\textbf{Hypothesis:}}
We focus on $F_2$, introduced in Section \ref{sec:interactions_example}.
For features $x_3$ to $x_6$, where there are non-zero interactions, we expect the different explanations to differ from the main effects. For features with no interactions ($x_7$ and $x_8$) or where the total effect is the same as the main effect ($x_1$ and $x_2$), we do not expect the explanations to differ significantly from the main effects.

\paragraph{\textbf{Setup:}} Similar to the previous section, we simulate 50,000 samples and then train a 2H-512,512 neural net to predict $F_2$.
We expect the interaction terms of $F_2$ to be absorbed by the different explanation methods. As a comparison, we plot marginal plots that attribute an interaction term to all features that made up that interaction.

\begin{figure}
\centering
\begin{tabular}{cccc}
\hspace*{-0.2in}
\begin{turn}{90}\hspace{1.3cm}  \scriptsize{$ h_i(x_i)$}\end{turn}
\includegraphics[width=0.32\linewidth]{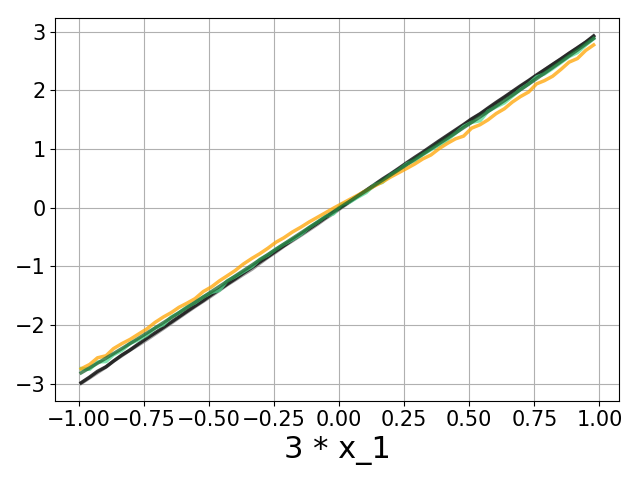} & \hspace*{-0.15in}
\includegraphics[width=0.32\linewidth]{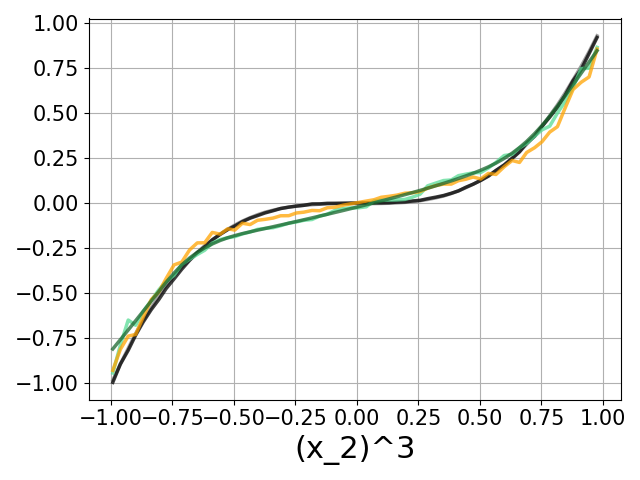} & \hspace*{-0.15in}
\includegraphics[width=0.32\linewidth]{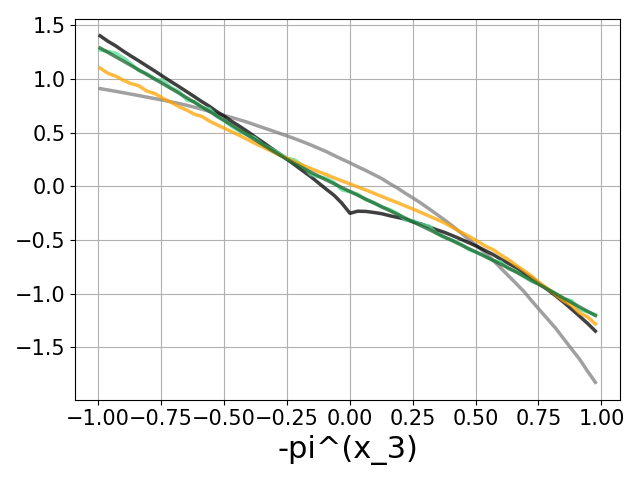} \\
\hspace*{-0.2in}
\begin{turn}{90}\hspace{1.3cm}  \scriptsize{$ h_i(x_i)$}\end{turn}
\includegraphics[width=0.32\linewidth]{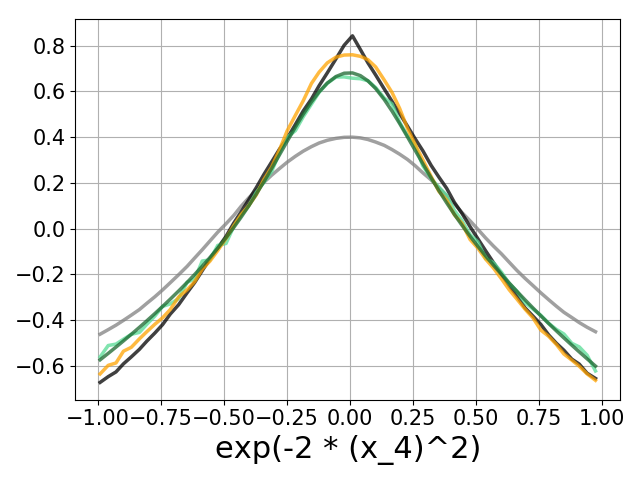} & \hspace*{-0.15in}
\includegraphics[width=0.32\linewidth]{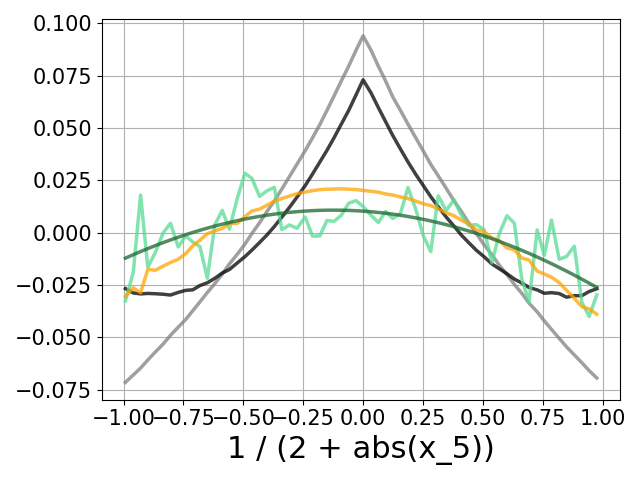} & \hspace*{-0.15in}
\includegraphics[width=0.32\linewidth]{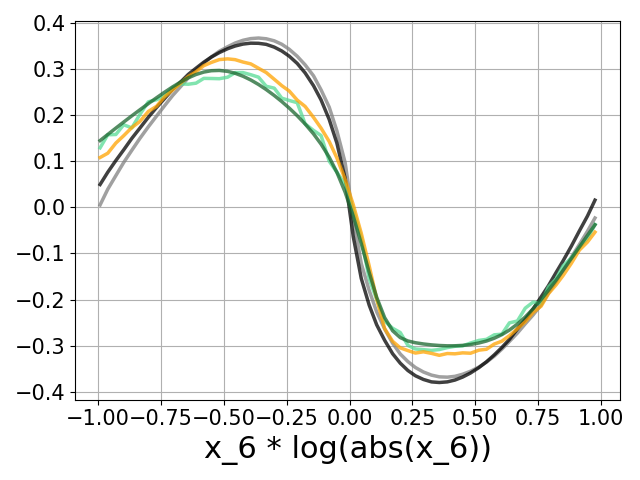} \\
\hspace*{-0.2in}
\begin{turn}{90}\hspace{1.3cm}  \scriptsize{$ h_i(x_i)$}\end{turn}
\includegraphics[width=0.32\linewidth]{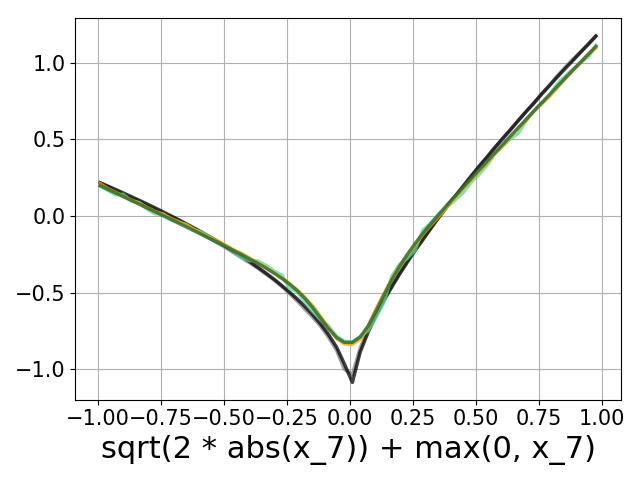} & \hspace*{-0.15in}
\includegraphics[width=0.32\linewidth]{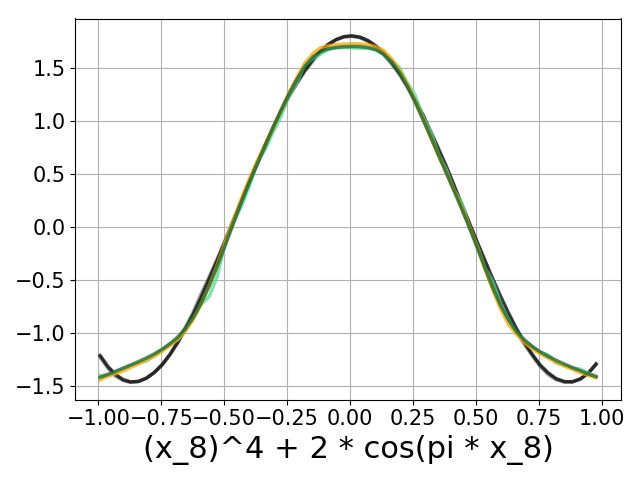} & \hspace*{-0.15in}
\includegraphics[width=0.32\linewidth]{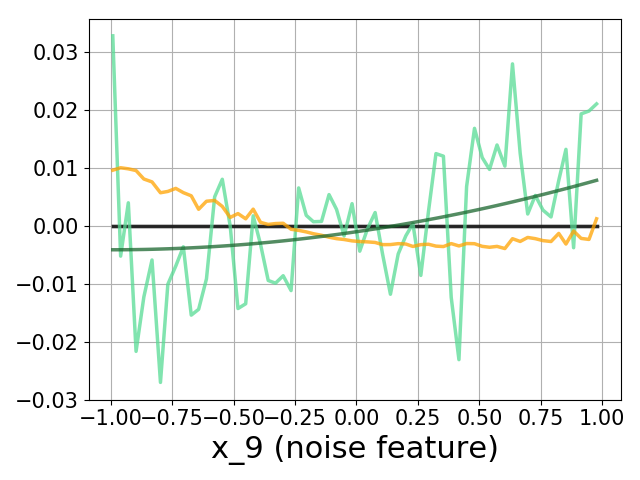}
\end{tabular}
\caption{For all features of function $F_2$: 
\textcolor{gray}{Main effect}, \textbf{Total effect}, \textcolor{green}{SAT explanation of complex black-box model (2H-512,512 neural net)},  \textcolor{greenmatplotlib}{PD explanation of the same 2H model}, and \textcolor{orange}{gSHAP explanation of the same 2H model}. Note that not all features are in the same scale, and that the scale of features $x_5$ and $x_9$ (a noise feature) are significantly smaller than the remaining features.}
\vspace{-0.5cm}
\label{fig:synthetic_2_explanations}
\end{figure}

\paragraph{\textbf{Results:}}
Figure \ref{fig:synthetic_2_explanations} shows the main effects (in \textcolor{gray}{gray}) and total effects (in \textbf{black}) for all the features.`

Figure \ref{fig:synthetic_2_explanations} also shows the explanations generated by the different methods. In general, the explanation have relatively similar shapes, although \cSAT and \cgSHAP tend to be less smooth than \cPD. This is not completely surprising, as \cSAT relies on tree learners, which do not lead to smooth transitions, and \cgSHAP uses less points than \cPD to compute the empirical contribution, since the \cgSHAP explanation is based on the conditional distribution while \cPD is based on the marginal. 

For features involved in interactions with non-zero expected contributions (\ie features $x_3$ to $x_6$), explanation methods absorb the contributions as expected. This is particularly clear for features $x_3$, $x_4$, and $x_5$, where the explanations look significantly more similar to the expected contribution of the interactions than to the mains. Out of the explanation methods, \cgSHAP is the most similar to the expected behavior, but all explanations are relatively similar.

\paragraph{\textbf{Implications:}}
Additive explanations are able to partially absorb the effect of interactions, as evidenced by the plots of features $x_3$ to $x_6$. However, the allocation of these effects is slightly different for each explanation method. Although we can use total effect as a point of reference, all allocations are valid, and it is not possible to quantify which characterization is better.

\subsection{How Does Correlation Between Features Affect Additive Explanations?}
\paragraph{\textbf{Hypothesis:}}
Explanations can be affected by feature correlation in non-obvious ways, for example by one feature absorbing, partially or totally, the importance of a different, correlated feature.

\paragraph{\textbf{Setup:}}
To illustrate the impact of correlation between features on additive explanations, we ran a controlled experiment, modifying the relationship between a feature and the label, but holding the relationship between another correlated feature and the label constant. 
For this task we use Bikeshare, an UCI data set where the goal is to predict bike usage given 12 features such as ``Season'', ``Day of the week'', ``Temperature'', ``Feeling temperature'', or ``Humidity''.

\begin{figure}
\centering
\begin{tabular}{cccc}
\hspace*{-0.2in}
\begin{turn}{90}\hspace{3cm}  \scriptsize{$h_i(x_i)$}\end{turn}
\includegraphics[width=0.5\linewidth]{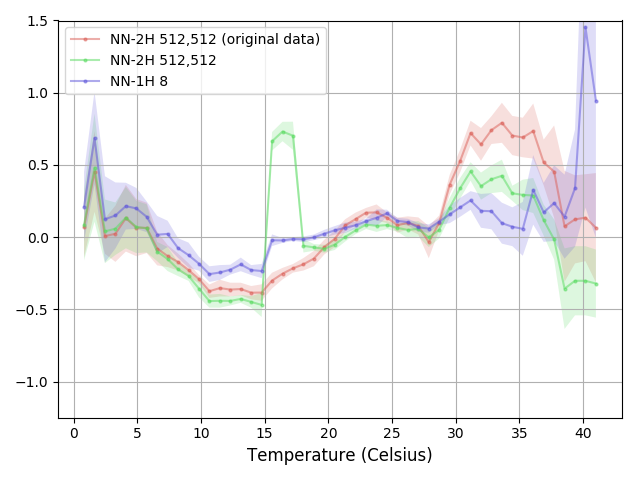} & \hspace*{-0.15in}
\includegraphics[width=0.5\linewidth]{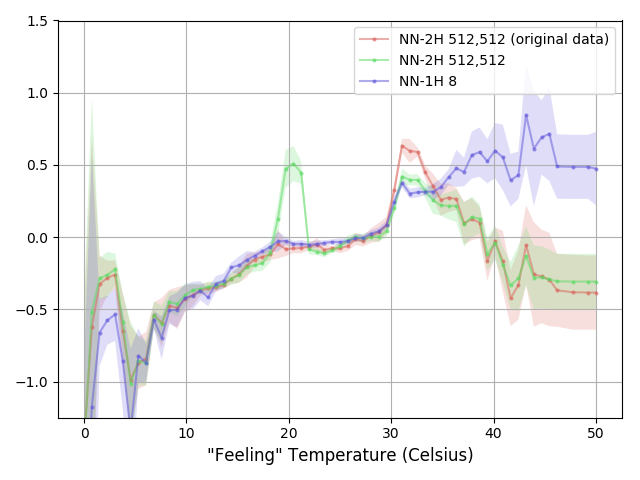} \\
 \multicolumn{2}{c}{\includegraphics[width=0.5\linewidth,trim={0cm 0cm 0 0.25cm},clip]{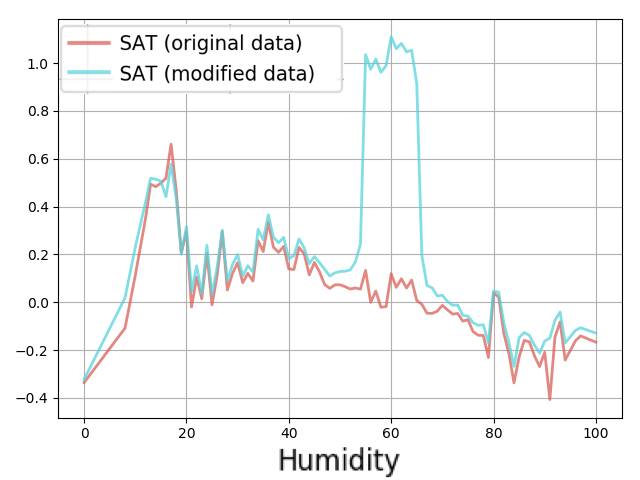}}
\end{tabular}
\caption{Additive explanations of two correlated features before and after label modification experiment. When the labels of certain values of temperature, a feature that is correlated with ``Feeling'' temperature, are modified, additive explanations of \textbf{both} features, not just the temperature feature, are impacted}
\label{fig:chapter_03_gae:bikeshare_temperature}
\end{figure}

For points with temperature between 15 and 18, we doubled the value of the label (the number of rented bikes) without making any modifications to the Feeling temperature feature, which is highly correlated with Temperature (Pearson correlation: 0.988). Separately, for points with humidify between 55 and 65, we increased the value of the label by 1.0 without making any modifications to other features. Humidity has low correlation with other features (the feature with the largest Pearson correlation is Windspeed: 0.346). In both cases, we retrained a 2H neural net on the modified data, and learned an additive explanation on the modified data. Figure \ref{fig:chapter_03_gae:bikeshare_temperature} displays the additive explanations trained on the original and modified data. For comparison, we also include the additive explanations for a 1H neural net trained on the modified data. 

\paragraph{\textbf{Results:}} Ideally, the additive explanations on the neural net trained on the modified data should be almost identical to the additive explanations on the neural net trained on the original data, except in that particular range of the temperature and humidity features, where we should see an abrupt bump corresponding to the controlled increase that we made. However, we see that a bump also occurs for the Feeling temperature feature, which is highly correlated with temperature.

\paragraph{\textbf{Implications:}} When features are correlated, additive explanations of correlated features distributed effects across them. This implies that the analysis of correlated features cannot be done independently. This can become a challenge when the number of correlated features is large.

\subsection{How Accurate are Additive Explanations on Non-Additive Models?}  
\label{sec:chapter_03_gae:comparing}

\paragraph{\textbf{Hypothesis:}}
A gap in accuracy between explanations and the original models is expected. Distilled explanations should generally be more accurate than non-distilled explanations.

\begin{table}
\centering
\begin{tabular}{l r l l l l l l}
  \toprule
  & & & & \multicolumn{3}{c}{\textbf{Performance}}\\
  \cmidrule(lr){5-7}
   \textbf{Data} & $n$ & $p$ & \textbf{Type} & & 1H & 2H\\ 
  \midrule
   Bikeshare & 17,000 & 12 & Regression & RMSE & 0.60 & 0.38\\
   Loan & 42,506 & 22 & Regression & RMSE & 2.71 & 1.91 \\
   Magic & 19,000 & 10 & Classification & AUC & 92.52 & 94.06\\
   Pneumonia & 14,199 & 46 & Classification & AUC & 81.81 & 82.18\\
   FICO & 9,861 & 24 & Classification & AUC & 79.08 & 79.37\\
   \bottomrule
\end{tabular}
\caption[Performance of black-box models]{Performance of black-box models. For RMSE, lower is better. For AUC, higher is better.}
\label{table:chapter_03_gae:data}
\end{table}

\paragraph{\textbf{Setup:}} We selected five data sets: two UCI data sets (Bikeshare and Magic), a Loan risk scoring data set from an online lending company \cite{lendingclub}, the 2018 FICO Explainable ML Challenge's credit data set \cite{fico2018}, and the pneumonia data set analyzed by \cite{caruana2015intelligible}. 
A 2H-512,512 neural net is used as the main black-box model. 
We quantitatively compare the accuracy of different types of global explanations on these data sets: distilled vs non-distilled explanations (\eg SAT vs gSHAP) and additive vs non-additive explanations (\eg SAT vs SAT+pairs or decision trees). We also compare the effect of explicitly modeling interactions in the explanation.

\begin{table*}[ht!]
\centering
\small 
\begin{tabular}{lcccccccc}
  \toprule
  \textbf{Accuracy} & & & & Bikeshare  & Loan   & Magic  &  Pneumonia & FICO\\
  Global Explanation & Additive & Distilled &  Interactions & RMSE & RMSE &  AUC & AUC & AUC \\
  \midrule
   SAT&Yes&Yes &No & 0.98 $\pm$ 0.00 & 2.35 $\pm$ 0.01 & 90.75 $\pm$ 0.06 & 82.24 $\pm$ 0.05 & 79.42 $\pm$ 0.04 \\
  SAS&Yes&Yes & No & 0.98 $\pm$ 0.00  & 2.34 $\pm$ 0.00  &  90.58 $\pm$ 0.02  & 82.12 $\pm$ 0.04  & 79.51 $\pm$ 0.02\\
  Sparse Linear&Yes&Yes & No & 1.39 $\pm$ 0.00 & 3.45 $\pm$ 0.00 & 86.91 $\pm$ 0.01 & 82.06 $\pm$ 0.02 & 79.16 $\pm$ 0.01\\
  \midrule
    gGRAD&Yes& No  & No & 1.25 $\pm$ 0.00  & 6.04 $\pm$ 0.01  & 80.95 $\pm$ 0.13  & 81.88 $\pm$ 0.05  & 79.28 $\pm$ 0.02\\
  gSHAP&Yes& No & No & 1.02 $\pm$ 0.00  & 5.10 $\pm$ 0.01  & 88.98 $\pm$ 0.05  & 82.31 $\pm$ 0.03  & 79.36 $\pm$ 0.01 \\
  PD&Yes&No & No & 1.00 $\pm$ 0.00  & 4.31 $\pm$ 0.00  & 82.78 $\pm$ 0.00  & 82.15 $\pm$ 0.00  & 79.47 $\pm$ 0.00 \\
  \midrule
    Decision Tree&No& Yes & Yes & 0.60 $\pm$ 0.01 & 2.66 $\pm$ 0.02 & 91.44 $\pm$ 0.29 & 79.38 $\pm$ 0.38 & 78.19 $\pm$ 0.03\\
  SAT+pairs&Yes& Yes & Yes & 0.60 $\pm$ 0.00 & 2.13 $\pm$ 0.01 &  90.75 $\pm$ 0.06 & 82.23 $\pm$ 0.06 & 79.44 $\pm$ 0.04\\
  \toprule
  
  \end{tabular}
  
\caption[Performance of global explanations for 2H black-box models]{Performance of global explanations for 2H black-box models. Performance measured in terms of RMSE for regression tasks and AUROC for classification tasks.}
\label{table:chapter_03_gae:all_derived_additive}
\end{table*}

\paragraph{\textbf{Results:}} 
First, for reference, Table \ref{table:chapter_03_gae:data} presents the accuracy of the 2H-512,512 black-box model, as well as the accuracy of a lower-capacity 1H-8 black-box model (provided for comparison purposes) and additional details about the datasets.
Then, Table \ref{table:chapter_03_gae:all_derived_additive} presents the accuracy of the different explanation methods on these datasets.
We draw several conclusions.
\begin{itemize}
    \item SAT and SAS yield similar accuracy, indicating that the particular choice of the base learner did not matter for these data sets. 
    \item Distilled explanations generally obtain better accuracy than non-distilled explanations. This is not surprising since distilled explanations were trained specifically to mimic the black-box model. However, on most datasets, there is still a gap between the accuracy of the explanations and the accuracy of the black box models.
    \item Non-linear models such as SAS, SAT, and decision trees outperform the sparse linear models on all datasets. However, sparse linear can be better than decision trees and competitive with SAS and SAT on some datasets (particularly Pneumonia and FICO), suggesting that complex models can easily overfit on some datasets with limited data (FICO and Pneumonia are the datasets with little data \emph{and} with the largest number of features, \cf Table \ref{table:chapter_03_gae:data}). These datasets are also the ones where the gap between the black-box models and the explanations is the smallest.
    \item When explicitly modeling interactions, the RMSE in  Bikeshare improves from 0.98 to 0.60. The Magic dataset is another example where the interactions brought by the decision tree make a significant difference. However, on some other datasets, decision trees are less accurate than other explanations, and, in the case of the Pneumonia dataset, even less accurate than a simple explanation based on a linear model.
\end{itemize}

\begin{figure}
\centering
\begin{tabular}{ccc}
\hspace*{-0.2in}
\begin{turn}{90}\hspace{2cm}  \scriptsize{$ h_i(x_i)$}\end{turn}
\includegraphics[width=0.45\linewidth]{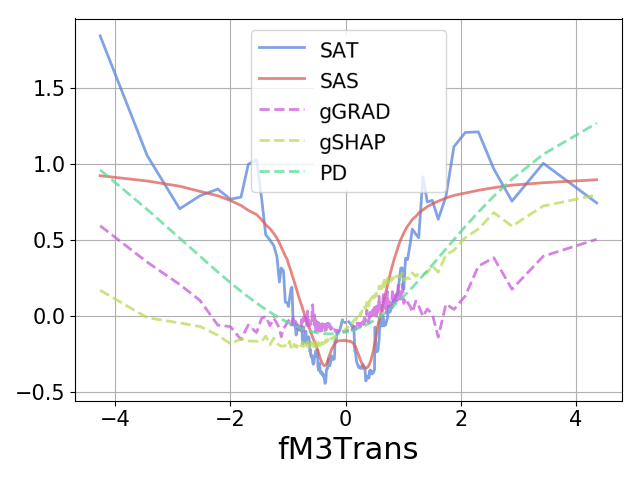} & \hspace*{0.15in}
\includegraphics[width=0.45\linewidth]{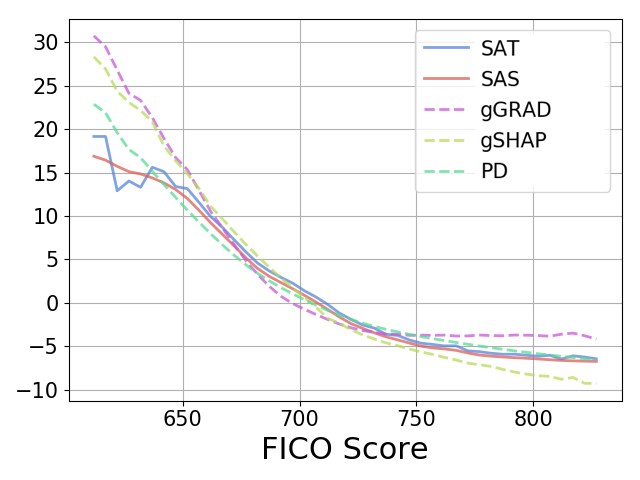} 
\end{tabular}
\caption[Example explanations for two datasets]{Example explanations of selected features for Magic data (left), and Loan data (right). SAT and SAS tend to agree.}
\label{fig:chapter_03_gae:samples_qualitative}
\end{figure}

Figure \ref{fig:chapter_03_gae:samples_qualitative} displays explanations of selected features for Magic and Loan. The explanations produced by PD tend to be much too smooth, which hurts its accuracy. Second, in all cases, trees and splines produce similar explanations and obtain equal or better accuracy and fidelity than the other methods. This is not surprising as the other methods are either local methods adapted to the global setting (gSHAP, gGRAD), or are global explanations that are not optimized to learn the teacher's predictions (PD). For reference, gSHAP when used as a local method (\ie individual SHAP values, not global explanations) achieved a lower RMSE of 0.37 compared to 1.02 on Bikeshare, and a lower RMSE of 1.99 compared to 5.10 on Loan, which is comparable to its 2H teacher's RMSE on test data (Table \ref{table:chapter_03_gae:data}). Hence, methods such as gSHAP excel at local explanations and should be used for those, but, to produce global explanations, global model distillation methods optimized to learn the teacher's predictions 
perform better.

\paragraph{\textbf{Implications:}}
Distilled explanations tend to be more accurate than non-distilled explanations. When the goal is to preserve accuracy, global explanations based on distillation are a better choice. Explicitly capturing interactions in the explanations may be unnecessary on some datasets (\eg FICO) or even pernicious (Pneumonia) while being extremely important in others (\eg Bikeshare). 

\section{How Interpretable Are Additive Explanations?}
\label{sec:chapter_03_gae:userstudy}
Thus far, we have considered how different additive explanations differentially allocate interaction effects, as well as their performance compared to non-additive explanations. We now conduct a user study to evaluate the interpretability of additive explanations compared to non-additive explanations.

\paragraph{\textbf{Explanations selected.}} Given the 2H-512,512 neural net described in Section \ref{sec:experiments} on the Bikeshare dataset, we learn four different additive and non-additive explanations using the same distillation techniques: (i) SAT explanations, (ii) Decision tree (DT) explanations, (iii) sparse linear (SPARSE) explanations, and (iv) Subgroup rules (RULES) explanations\footnote{State-of-the-art rule lists \cite{letham2015interpretable,angelino2017learning} do not support regression, which is needed for distillation. We used a slightly older subgroup discovery algorithm \cite{vikamine} that supports regression but does not generate disjoint rules. This method only achieved reasonable results on Bikeshare.}. 
Denoting the ``complexity'' of the explanations by explanation-K, where the meaning of K depends on the underlying model\footnote{For decision trees K represents the depth, and a tree of depth 4 would be denoted as DT-4. For sparse rules, K represents the number of rules, and  a group of 5 rules would be denoted as RULES-5. For SAT and SPARSE, K denotes the number of features to use. In the case of SPARSE, K is set indirectly by finding the regularization lambda parameter that produced the best accuracy on validation while also producing exactly K non-zero feature coefficients.}, we select the following explanations: SAT-5, DT-4, SAT-2, DT-2, RULES-5, and SPARSE-2. 

\begin{figure*}
\centering
\fbox{\includegraphics[width=0.8\linewidth]{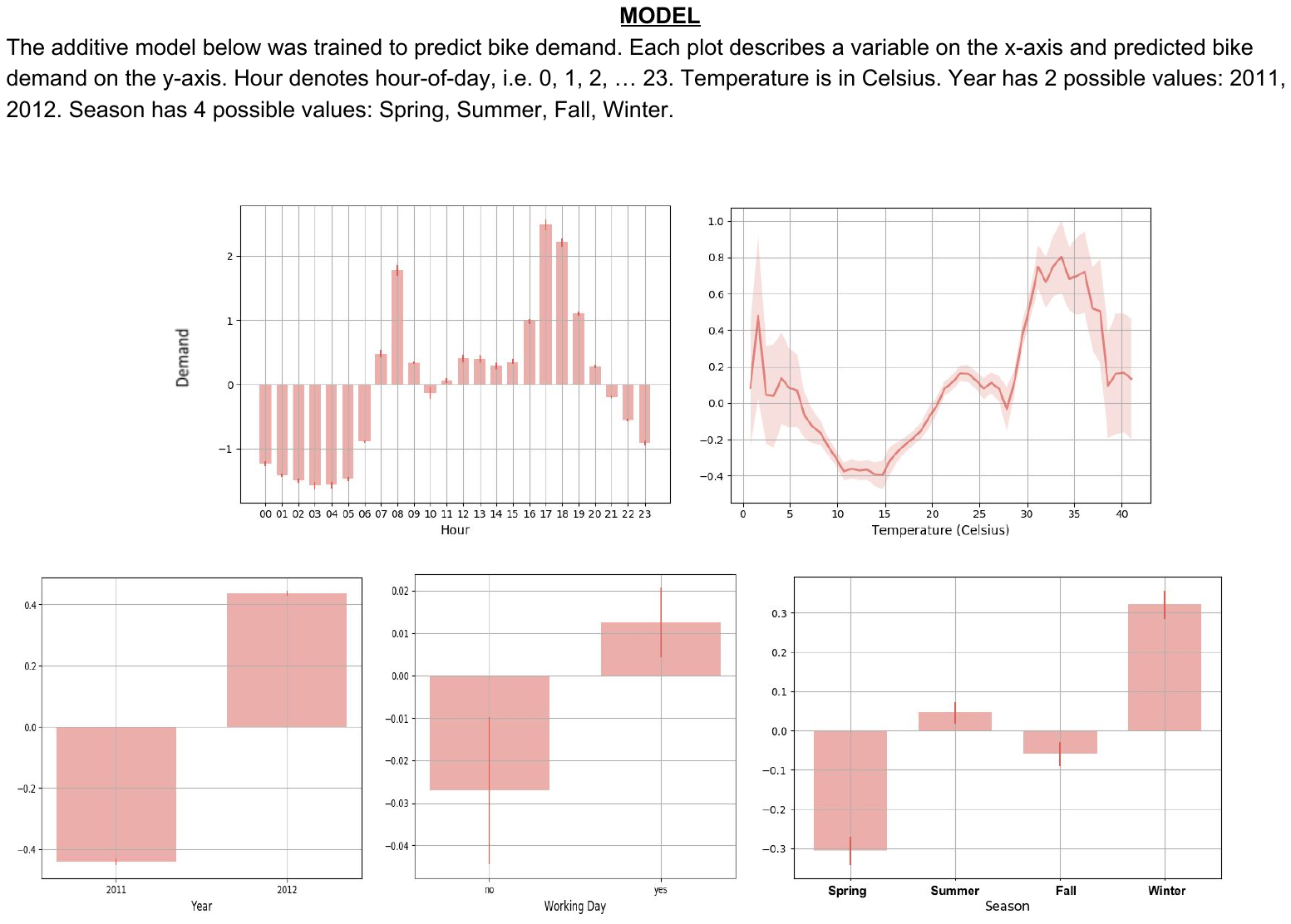}}

\vspace{1cm}

\fbox{\includegraphics[trim={0 9cm 0 0},clip,width=0.95\linewidth]{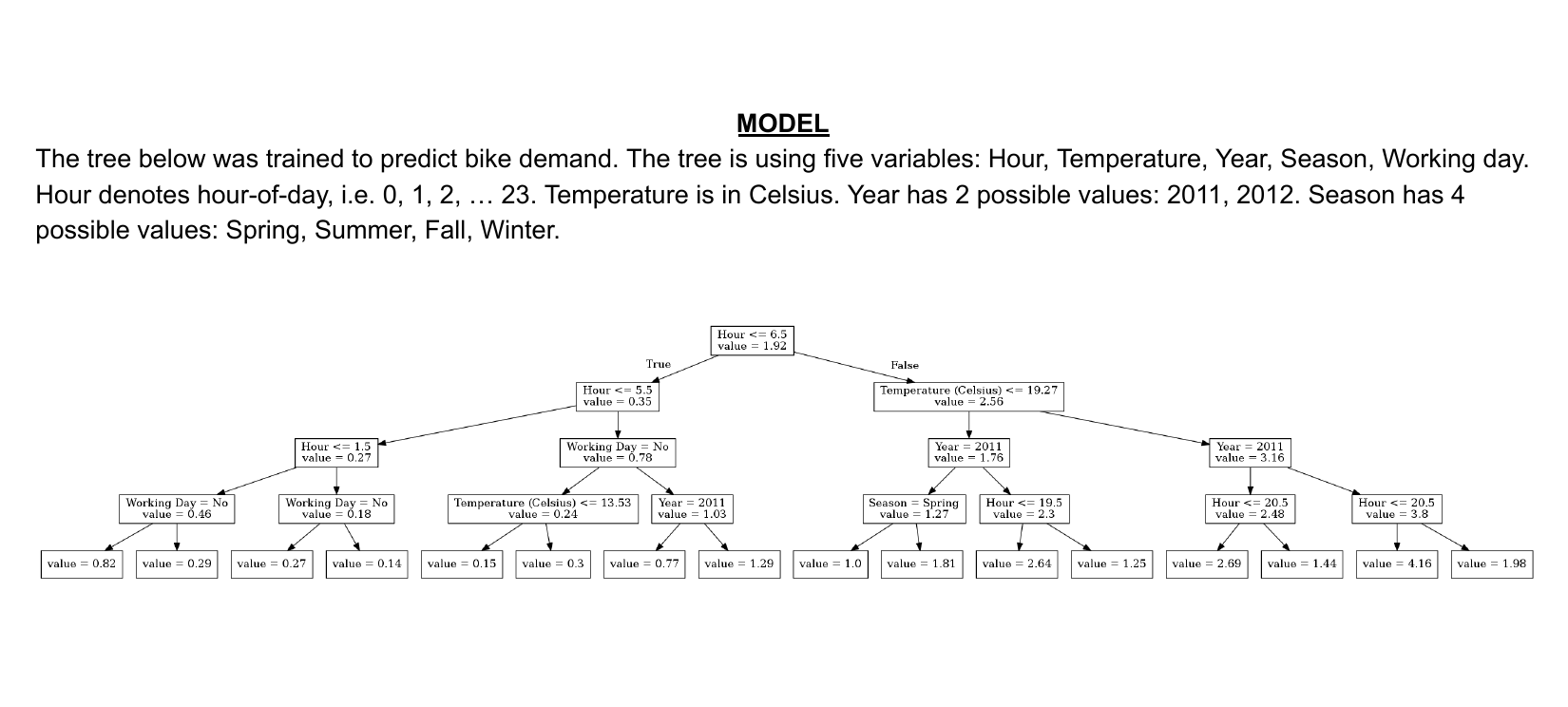}}
\caption{Explanations shown to SAT-5 and DT-4 subjects in user study}
\label{fig:chapter_03_gae:biggambigtree}
\end{figure*}

\begin{figure*}
\centering
\fbox{\includegraphics[trim={0 15cm 0 0},clip,width=0.95\linewidth]{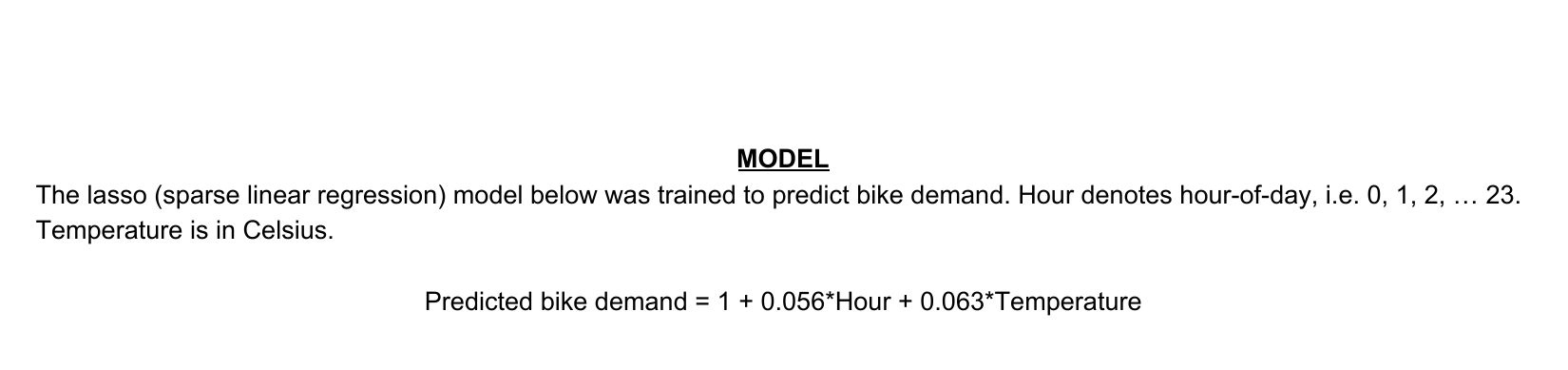}}

\vspace{1cm}

\fbox{\includegraphics[trim={0 4.5cm 0 0},clip,width=0.95\linewidth]{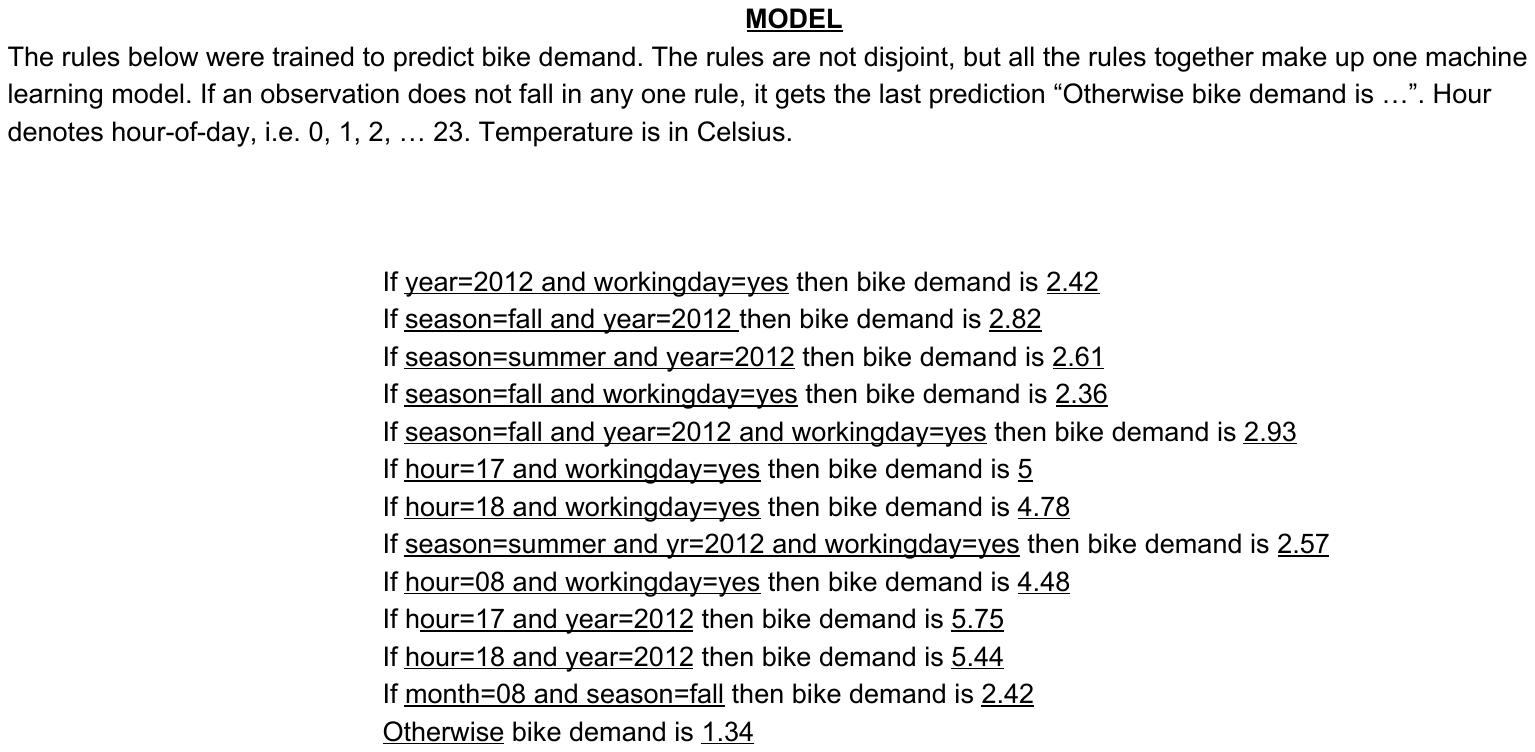}}
\caption{Explanations shown to SPARSE-2 and RULES-5 subjects in user study}
\label{fig:chapter_03_gae:sparserules}
\end{figure*}

\begin{figure}[htbp]
\centering
\includegraphics[width=\linewidth]{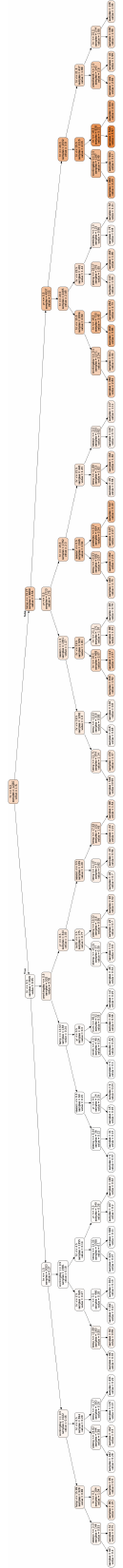}
\caption[Tree of depth 6 (64 leaves), the least deep tree that matched SAT's fidelity]{Tree of depth 6 (64 leaves), the least deep tree that matched SAT's fidelity. This uses the default tree visualizer in scikit-learn. Note that the tree is so large that it is hard to read even in digital form at the maximum publishing resolution.}
\label{fig:chapter_03_gae:hugetree}
\end{figure}

We selected DT-4 because that is the largest tree that is readable on letter-size paper, and that does not lag too far behind the depth 6 tree in accuracy: the DT-4 has an accuracy of 1.16 RMSE, compared to the 1.0 RMSE of DT-6 and the 0.98 RMSE of SAT. For reference, we show the DT-6 tree in Figure \ref{fig:chapter_03_gae:hugetree}. DT-6 is slightly more accurate than DT-4, but much harder to read.
DT-4 used five features: Hour, Temperature, Year, Working Day, Season. To make the explanations more comparable, we select those features for the SAT explanation, hence SAT-5, which has a comparable accuracy of 1.07 RMSE. Figure \ref{fig:chapter_03_gae:biggambigtree} displays the exact DT-4 and SAT-5 explanations shown to the subjects.
We also explored smaller and/or less accurate explanations to assess the trade-off between accuracy and explainability. We selected the top 2 features and built SAT-2 (1.14 RMSE), DT-2 (1.3 RMSE), and SPARSE-2 (1.54 RMSE) explanations. Last, we also considered a rules-based explanation, which is less accurate even with a larger number of features: RULES-5 achieves an RMSE of 1.47, similar to SPARSE-2. Figure \ref{fig:chapter_03_gae:sparserules} displays the exact SPARSE-2 and RULES-5 explanations shown to the subjects.

\paragraph{\textbf{Study design.}} 50 subjects were recruited to participate in the study. These subjects -- STEM PhD students, or college-educated individuals who had taken a machine learning course -- were familiar with concepts such as if-then-else structures (for trees and rule lists), reading scatterplots (for SAT), and interpreting equations (for sparse linear models). Each subject only used one explanation model (between-subject design) to answer a set of questions 
covering common inferential and comprehension tasks on machine learning models: (1) Rank features by importance; (2) Describe relationship between a feature and the prediction; (3) Determine how the prediction changes when a feature changes value; (4) Detect an error in the data, captured by the model. The exact questions were: 

\begin{enumerate}
    \item What is the most important variable for predicting bike demand?
\item Rank all the variables from most important to least important for predicting bike demand.
\item Describe the relationship between the variable Hour and predicted bike demand. 
\item What are variables for which the relationship between the variables and predicted bike demand is positive? 
\item The Hour is 11. When Temperature increases from 15 to 20, how does predicted bike demand change?
\item There is one error in the data. Any idea where it might be? ``Cannot find the error'' is an ok answer. 
\end{enumerate}

In the first stage, 24 of 50 subjects were randomly assigned to see output from DT-4 or SAT-5. In the second stage, we experimented with smaller versions of trees and SAT using only the two most important features, Hour and Temperature. 14 of 50 subjects were randomly assigned to see output from SAT-2 or DT-2. In the last stage, the remaining 12 subjects were randomly assigned to see output from one of the two worst performing explanations (in terms of accuracy): sparse linear models (SPARSE-2) and subgroup-rules (RULES-5).

\paragraph{\textbf{Can humans understand and use explanations?}} From the absolute magnitude of the SAT feature explanations as well as Gini feature importance metrics for the tree, we determined the ``ground truth'' feature importance ranking (in decreasing order): Hour, Temperature, Year, Season, Working Day. More SAT-5 than DT-4 subjects were able to rank the top 2 and all features correctly (75\% vs. 58\%, see Table \ref{table:chapter_03_gae:study_big}). 

\begin{table}
\footnotesize
\bgroup
\def\arraystretch{1.15}
\begin{tabular}{lcccccc}
  \toprule
  & \multicolumn{2}{c}{First stage (n=24)} & \multicolumn{2}{c}{Second stage (n=14)} & 
  \multicolumn{2}{c}{Third stage (n=12)} \\
   \cmidrule(lr){2-3} \cmidrule(lr){4-5} \cmidrule(lr){6-7} 
   \textbf{Task} & \textbf{SAT-5}  & \textbf{DT-4} & \textbf{SAT-2} & \textbf{DT-2}  & \textbf{SPARSE-2}  & \textbf{RULES-5}  \\ 
  \midrule
  Ranked correctly top 2 features & 75\% & 58\% & 100\% & 85.7\% & 83.3\% & 0\%\\
  Ranked correctly all (5) features & 45\% & 0\% & N/A & N/A & N/A & 0\%\\
NDCG between human ranking of top 5 features  & \multirow{2}{*}{0.94 $\pm$ 0.13} & \multirow{2}{*}{0.89 $\pm$ 0.11} & \multirow{2}{*}{N/A} & \multirow{2}{*}{N/A} & \multirow{2}{*}{N/A}   & \multirow{2}{*}{0.27 $\pm$ 0.11}\\
\hspace{0.3cm}  and ground-truth feature importance \\
 \midrule
  Described increased demand & \multirow{2}{*}{42\%} & \multirow{2}{*}{0\%} & \multirow{2}{*}{29\%} & \multirow{2}{*}{0\%} & \multirow{2}{*}{0\%} & \multirow{2}{*}{33\%}  \\
  \hspace{0.3cm} during rush hour   \\
  Described increased demand &  \multirow{2}{*}{33\%} &  \multirow{2}{*}{0\%} &  \multirow{2}{*}{29\%} &  \multirow{2}{*}{0\%} & \multirow{2}{*}{0\%} & \multirow{2}{*}{33\%}\\
  \hspace{0.3cm} during mornings and afternoons \\
  \midrule
  Compute change in prediction & \multirow{2}{*}{33\%} & \multirow{2}{*}{25\%} & \multirow{2}{*}{14\%} & \multirow{2}{*}{100\%} & \multirow{2}{*}{83\%} & \multirow{2}{*}{0\%}\\
  \hspace{0.3cm} when feature changes  \\
  \midrule
    Caught data error & 33\% & 8\% & N/A & N/A & N/A & 0\%\\
    \midrule
  Time taken (minutes) & 11.7 $\pm$ 5.8  & 17.5 $\pm$ 14.8 & 7.2 $\pm$ 3.2  & 6.2 $\pm$ 2.2 & 5.2 $\pm$ 3.1 & 14.9 $\pm$ 8.4 \\
   \bottomrule
\end{tabular}
\egroup

\caption[Quantitative results from user study on expert users]{Quantitative results from user study. Since SAT-2, DT-2, and SPARSE-2 only had two features, the task to rank five features does not apply. Since the data error only appeared in the output of SAT-5, DT-4, and RULES-5, the other subjects could not have caught the error.}
\label{table:chapter_03_gae:study_big}
\end{table}

When ranking all 5 features, 0\% of the DT-4 and RULES-5 subjects were able to predict the right order, while 45\% of the SAT-5 subjects correctly predicted the order of the 5 features, showing that ranking feature importance for trees is actually a very hard task.
The most common mistake made by DT-4 subjects (42\% of subjects) was to invert the ranking of the last two features, Season and Working Day, perhaps because Working Day's first appearance in the tree (in terms of depth) was before Season's first appearance (Figure \ref{fig:chapter_03_gae:biggambigtree} bottom).
We also evaluate the normalized discounted cumulative gain (NDCG) between the ground truth feature importance and the user prediction, where we give relevance scores to the feature in decreasing order (\ie, for 5 features, the most important feature has a relevance score of 5, the second most important 4, etc). This gives us an idea of \emph{how well} the features were ranked, even if the ranking is not perfect. We see how SAT-5 obtains a better score than DT-4, consistent with the previous analysis. RULES-5 obtains a significant lower score.

When asked to describe, in free text, the relationship between the variable Hour and the label, one SAT-5 subject wrote: 
\begingroup
\addtolength\leftmargini{-0.2in}
\begin{quote}
\emph{There are increases in demand during two periods of commuting hours: morning commute (\eg 7-9 am) and evening commute
(\eg 4-7 pm). Demand is flat during working hours and predicted to be especially low overnight,}
\end{quote}
\endgroup
\vspace{-0.15cm}
whereas DT-4 subjects' answers were not as expressive, \eg: 
\vspace{-0.05cm}
\begingroup
\addtolength\leftmargini{-0.2in}
\begin{quote}
\emph{Demand is less for early hours, then goes up until afternoon/evening, then goes down again.}
\end{quote}
\endgroup

75\% of SAT-5 subjects detected and described the peak patterns in the mornings and late afternoons, and 42\% of them explicitly mentioned commuting or rush hour in their description. On the other hand, none of the DT-4 subjects discovered this pattern on the tree: most (58\%) described a concave pattern (low and increasing  during the night/morning, high in the afternoon, decreasing in the evening) or a \emph{positively correlated} relation (42\%). 
Similarly, more SAT-5 subjects were able to precisely compute the change in prediction when temperature changed in value, and detect the error in the data -- that spring had lower bike demand whereas winter had high bike demand (bottom right feature in Figure \ref{fig:chapter_03_gae:biggambigtree} top).

\begin{figure*}
\centering
\includegraphics[height=5cm]{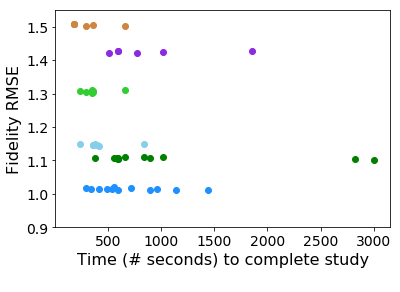}\hspace{0.15cm}
\includegraphics[height=5cm]{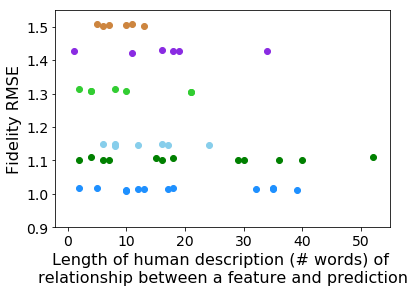}
\includegraphics[height=5cm]{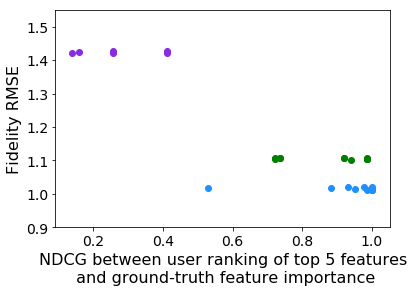}
\caption[User study metrics, as proxies for interpretability, by fidelity for different explanations]{User study metrics, as proxies for interpretability, by fidelity (RMSE) for different explanations. Each point is an individual user in the user study. The metrics are time needed to finish the study (top lefte), length of the description (top right), and the NDCG of the ranked features (bottom). Key:  \gamfive, \dtfive, \gamtwo, \dttwo, \rulesfive, \sparsetwo}
\label{fig:chapter_03_gae:study_model}
\end{figure*}

\paragraph{\textbf{How do tree depth and number of features affect human performance?}}
We also experimented with simpler explanations, SAT-2 and DT-2, that used  only the two most important features, Hour and Temperature.
As the explanations are simpler, some of the tasks become easier. For example, SAT-2 subjects predict the order of the top 2 features 100\% of the time (vs 75\% for SAT-5), and DT-2 subjects, 85\% of the time (vs 58\% for DT-4). The most interesting change is in the percentage of subjects able to compute the change in prediction after changing a feature: only 25\% for DT-4, compared to 100\% for DT-2. Reducing the complexity of the explanation made using it easier, \emph{at the price of reducing the accuracy of the explanation}. 
Another important aspect is the time needed to perform the tasks: increasing the number of features from 2 to 5 increases the time needed by the subjects to finish the study by 60\% for the SAT explanation, but increases it by 166\% for the DT explanation, that is, interpreting a tree becomes much more costly as the tree becomes deeper (and more accurate), and, in general, subjects make more mistakes. SAT scales up more gracefully.

\paragraph{\textbf{Remaining explanations: subgroup-rules and sparse linear models.}}
These explanations were the least accurate and faithful.
We found that human subjects can easily read the (few) weights of SPARSE-2, establish feature importance, and compute prediction changes, and do so quickly -- at $5.1$ minutes on average, this was the fastest explanation to interpret. However, the explanation is highly constrained and hid interesting patterns. For example, 100\% of the subjects described the relation between demand and hour as increasing, and 83\% predicted the exact amount of increase, but none were able to provide insights like the ones provided by SAT-5 and DT-4 subjects.

RULES-5 was the second hardest explanation to interpret based on mean time required to answer the questions: 14.9 minutes. Understanding non-disjoint rules appears to be hard: none of the subjects correctly predicted the feature importance order, even for just two features; none were able to compute exactly the change in prediction when feature value changes, and none were able to find the data error. The rules in RULES-5 are not disjoint because we could not find a regression implementation of disjoint rules. However, 66\% of the subjects discovered the peak during rush hour, as that appeared explicitly in some rules, \eg ``If hour=17 and workingday=yes then bike demand is 5''.

\paragraph{\textbf{Fidelity vs. interpretability.}}
Figure \ref{fig:chapter_03_gae:study_model} presents detailed results for individual users by model. On the left is the time needed to finish the study (left). In the center is the length of the user's written description of the relationship between a feature and model predictions. On the right is the NDCG rank loss of user ranking of feature importance compared to ground-truth feature importance. All of these metrics can be considered interpretability metrics, when defining interpretability as grounded in human tasks \cite{doshivelez2017towards}. On the y-axis is fidelity (RMSE), \ie, how similar is the explanation to the 2H model prediction.

The plots show that there is a trade-off between fidelity and interpretbility (as measured by time to complete, description length, and NDCG of feature rankings), but not all methods behave similarly. In general, the SPARSE-2 explanation is easy to understand (users typically finish the study rapidly), but fidelity is poor and it leads to short descriptions. On the other hand, SAT-5 and DT-4 have much better fidelity and lead to more varied descriptions, but also took longer to understand. DT-2 was faster to complete than DT-4, but the fidelity is lower and the descriptions shorter.  RULES-5 is better than SPARSE-2, but not as good as SAT-5 or DT-4. SAT-5 offers a reasonable trade-off, being both faithful and relatively easy to understand, while also leading to rich descriptions for many users.
 
 To summarize, global additive explanations: (1) allowed humans to perform better (than decision trees, sparse linear models, and rules) at ranking feature importance, pointing out patterns between certain feature values and predictions, and catching a data error; (2) Additive explanations were also faster to understand than big decision trees; (3) Very small decision trees and sparse linear models had the edge in calculating how predictions change when feature values change, but were much less faithful and accurate.

\section{Conclusions}
\label{sec:conclusions}
In this work we studied 
how additive explanations behave when explaining black-box models with non-additive components. We showed that in this case, there is no one unique additive explanation, with different explanation methods  characterize these non-additive components in different ways. We quantitatively compared different additive explanations on several regression and classification tasks, finding that distilled explanations are generally the most accurate additive explanations.
Although  non-additive explanations that explicitly model interactions tend to be more accurate, we found, through a user study, that machine learning practitioners were able to leverage additive explanations better. When deciding which explanation to use for a black-box model, trustworthiness, accuracy, and interpretability should all be considered. 

\section{Acknowledgements}
\label{sec:acknowledgements}
We thank Julius Adebayo for helpful discussion.

\section{Declarations}
\noindent \textbf{Funding} - Giles Hooker was supported by NSF grant DMS-1712554 and DEB-1933497.

\noindent \textbf{Conflicts of interest/Competing interests} - The authors declare that there is no conflict of interest.

\noindent \textbf{Ethics approval} - Not applicable.

\noindent  \textbf{Consent to participate} - User study subjects consented to participate in the user study.

\noindent \textbf{Consent for publication} - Not applicable.

\noindent  \textbf{Availability of data and material} - Publicly available datasets used in this paper can be found at the websites of: UCI (Bikeshare, Magic), Kaggle (Lending Club), FICO. Whenever available, links are cited in this paper's references.

\noindent \textbf{Code availability} - Publicly-available code to help other researchers replicate the work can be found at: \url{https://github.com/shftan/distilled_additive_explanations}.

\noindent \textbf{Authors' contributions} - The authors contributed to this paper in the following manner: Sarah Tan designed, executed, and analyzed the experiments and wrote the paper. Giles Hooker formulated mathematics in the paper and wrote the paper. Paul Koch wrote software used by the experiments. Albert Gordo executed experiments and wrote the paper. Rich Caruana analyzed experiments and wrote the paper.

\bibliographystyle{ACM-Reference-Format}
\bibliography{main}

\end{document}